\def\year{2020}\relax
\documentclass[letterpaper]{article} 
\usepackage{aaai20}  
\usepackage{times}  
\usepackage{helvet}  
\usepackage{courier}  
\usepackage{url}  
\usepackage{graphicx}  
\usepackage{amsmath}
\usepackage{dcolumn}
\usepackage{kotex}
\usepackage{color}
\usepackage{hhline}
\usepackage{verbatim}
\usepackage{multirow}
\usepackage{amssymb}
\usepackage{rotating}
\usepackage{subfigure}
\usepackage{enumitem}
\usepackage[dvipsnames]{xcolor}
\usepackage[T1]{fontenc}

\begin{document}
%

\title{Lightweight and Robust Representation of Economic Scales\\ from Satellite Imagery}

\author{
  Sungwon Han\thanks{The first three authors contributed equally.}\textsuperscript{\rm 1,3}~~~
  Donghyun Ahn\footnotemark[1]\textsuperscript{\rm 1,3}~~~ 
  Hyunji Cha\footnotemark[1]\textsuperscript{\rm 1,3}\\
  \bf \Large Jeasurk Yang\textsuperscript{\rm 2,3}~~~  
  \bf \Large Sungwon Park \textsuperscript{\rm 1,3}~~~
  \bf \Large Meeyoung Cha\textsuperscript{\rm 3,1}\\
  \textsuperscript{\rm 1}School of Computing, KAIST, South Korea\\ 
  \textsuperscript{\rm 2}The Institute for Korean Regional Studies, SNU, South Korea \\
  \textsuperscript{\rm 3}Data Science Group, IBS, South Korea\\
  \{lion4151, segaukwa, hyunji3190, psw0416, meeyoungcha\}@kaist.ac.kr~~
  hatomi91@snu.ac.kr 
}
\maketitle
\begin{abstract}
Satellite imagery has long been an attractive data source providing a wealth of information regarding human-inhabited areas. While high-resolution satellite images are rapidly becoming available, limited studies have focused on how to extract meaningful information regarding human habitation patterns and economic scales from such data. We present READ, a new approach for obtaining essential spatial representation for any given district from high-resolution satellite imagery based on deep neural networks. Our method combines transfer learning and embedded statistics to efficiently learn the critical spatial characteristics of arbitrary size areas and represent such characteristics in a fixed-length vector with minimal information loss. Even with a small set of labels, READ can distinguish subtle differences between rural and urban areas and infer the degree of urbanization. An extensive evaluation demonstrates that the model outperforms state-of-the-art models in predicting economic scales, such as the population density in South Korea ($R^2$=0.9617), and shows a high use potential in developing countries where district-level economic scales are unknown.
\end{abstract}

\section{Introduction}
Satellite images captured through remote sensing technology provide useful information regarding human activities and land covers without having to visit the region. Analyzing satellite images provides new insights for establishing policies and for understanding population-level behaviors in districts and broader areas. Another advantage of satellite imagery is its ability to search historical data, which can reveal the temporal dynamics of specific locations. Several techniques have been proposed to analyze satellite images.

\begin{figure}[t!]
    \centering
    \includegraphics[width=0.88\columnwidth]{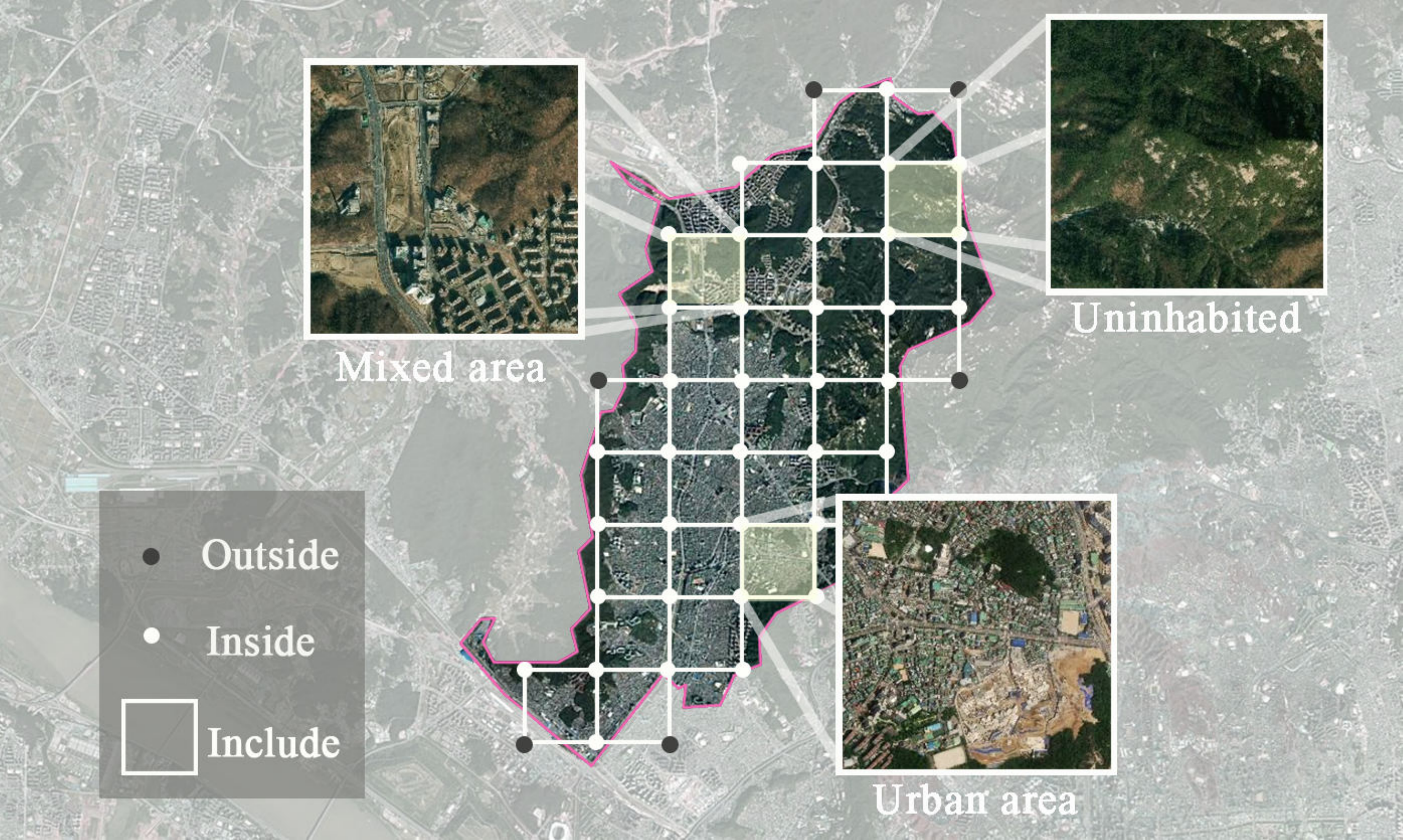}
    \caption{Methods for selecting satellite imagery of target districts. Unlike the previous methods~\cite{jean2016combining}, our neural network model can utilize all image tiles belonging to the target. At least three points should belong to the district. The mixed area is a mixed landscape composed of urban and rural interaction.}
    \label{fig:sate_demo2}
\end{figure}

Many studies measure the critical factors of socioeconomic scales from satellite images, and the demand has been rapidly increasing. For example, demographic research represents a popular application area in which scholars have leveraged the light intensity of nighttime satellite images to extract economic activities~\cite{ghosh2013using,chaturvedi2011assessing,letu2011monitoring}. Nighttime satellite images have also been used to predict the population density of counties in the United States~\cite{sutton1997comparison}. Although nighttime images are subject to noise, such as saturation effects, they have shown to yield a meaningful linear relationship between brightness and various economic scales.

More recently, high-resolution daytime satellite imagery has emerged as an attractive data source that could reveal fine-grained information at an unprecedented scale. For example, transfer learning has been used to extract socioeconomic indicators to create an accurate poverty map of Uganda~\cite{xie2016transfer}. Such research has considerable implications for sustainable development growth in developing countries. A convolutional neural network (CNN) has been used to predict grid cell estimates of county-level population counts~\cite{robinson2017deep}. This work produced a high-resolution grid-shaped population map from satellite images at a 30 arc-second resolution ($\approx$ 1$km^{2}$). However, this method is not applicable outside of the United States because conventional statistics are produced based on districts that can be of any polygon shape rather than at the grid-level, which leads to a mismatch when attempting to use satellite images.

The method proposed in this paper is inspired by the mean-teacher model in semi-supervised learning~\cite{tarvainen2017mean} and the full-convolution CNN model with transfer learning capacity used to predict poverty~\cite{jean2016combining}. Our method, which is called `Representation Extraction over an Arbitrary District' (\textbf{READ}), utilizes daytime satellite image tiles whose three vertices belong entirely to the polygon representing each district. This method is depicted in Fig.~\ref{fig:sate_demo2}. As the picture demonstrates, a single district can contain vastly different land covers such as urban built-up, water, forest, etc. Our task is to learn these sophisticated spatial features of an arbitrarily shaped district based on high-resolution satellite images to produce a fixed-length representation of economic scales.

The model has considerable implications for monitoring the urbanization process not only in developing countries but also in developed countries. Cities are rapidly evolving; however, it is far slower to detect this process, often using decade-cycled demographic surveys. Notably, these surveys are even extraordinarily costly and time-consuming; our model offers a possible alternative with a reasonable budget and time requirement. Our method can measure diverse urban phenomena such as urban sprawl with improved temporal resolution.

READ is a lightweight method of measuring economic scales from high-resolution images. The learned features are robust to the size of the original labels. They are highly informative when estimating critical economic scales, such as population density, age, education, income, etc. We present a comprehensive evaluation of the model based on a rich set of data from one developed country, South Korea, and demonstrate its potential use in a developing country, Vietnam. The code is released at GitHub.\footnote{https://github.com/Sungwon-Han/READ}

\section{Related works}
Satellite images have become available for public use. These images allow the constant monitoring of the earth and reveal the detailed land cover during both nighttime and daytime, rendering them an excellent resource for the prediction of human activities. Nighttime satellite images are in a lower resolution and have been used to predict the gross domestic product~\cite{sutton2007estimation,chen2011using}, energy consumption~\cite{xie2016world,hu2019novel}, epidemic fluctuations~\cite{bharti2011explaining}, and regional economic productivity~\cite{doll2006mapping} at the national level. In contrast, high-resolution daytime satellite imagery can reveal detailed land appearance over smaller areas.

Scholars have applied daytime satellite images to a CNN-based model of the Visual Geometry Group 16 (VGG16) architecture to predict the population density~\cite{Simonyan15}. Other scholars~\cite{doupe2016equitable} used 250m-resolution Landsat-7 satellite imagery as input to estimate the population density of Kenya. A regression model was used in the study, while a subsequent study~\cite{robinson2017deep} used a classification model to determine the electricity use levels of the population using US Census Summary Grids. Furthermore, This approach~\cite{vogel2018detecting} is used to integrate nighttime lighting data with 30m-resolution Landsat-8 satellite images and apply machine learning techniques to detect urban markets that help capture real economic activity. The latest study conducted by Facebook Research and Columbia University released an online high-resolution population density map of Africa and Asia~\cite{facebook2019data}. Their basic idea is to perform a binary classification of each satellite image that verifies the existence of a building and rankings the likelihood of an area having urban structures. While previous models had adapted VGG for training, this new model applied residual neural nets. This work contributed to the construction of the gridded population of the world version 4~\cite{ciesin2016}, which is known as the current state-of-the-art global population density map.

Satellite images have also been used for poverty prediction. At the national level, a previous study reported a linear relationship between nighttime lights and the percent of the population living on \$2 per day or less using the Defense Meteorological Satellite Program (DMSP) F-15 images~\cite{elvidge2009global}. Another study predicted the income level in India using developmental statistics~\cite{pandey2018multi}. These authors trained a convolutional network to predict parameters and income levels. However, the parameters utilized in this model are specific to the country analyzed.

Deep learning methods can predict demographic information through satellite images at the national level. The performance of models trained using supervised learning largely depends on the quality and number of the labeled data~\cite{Xiao_2015_CVPR}. This is because of the lack of labeled data, which makes it challenging to train deep neural networks. A recent study proposed a novel approach to address this deficiency by adopting two-staged transfer learning and testing a linear regression of fine-tuned feature extractors~\cite{jean2016combining}. However, this method is subject to noise because the satellite images had to be randomly chosen for each target area to reduce the computation overhead. Compared to this study, the current paper proposes to utilize all available images instead of sampling via its unique lightweight structure. Our approach also enables computation over smaller and flexible geographical boundaries.

\section{Model}

\begin{figure}[t!]
    \centering
    \includegraphics[width=1\columnwidth]{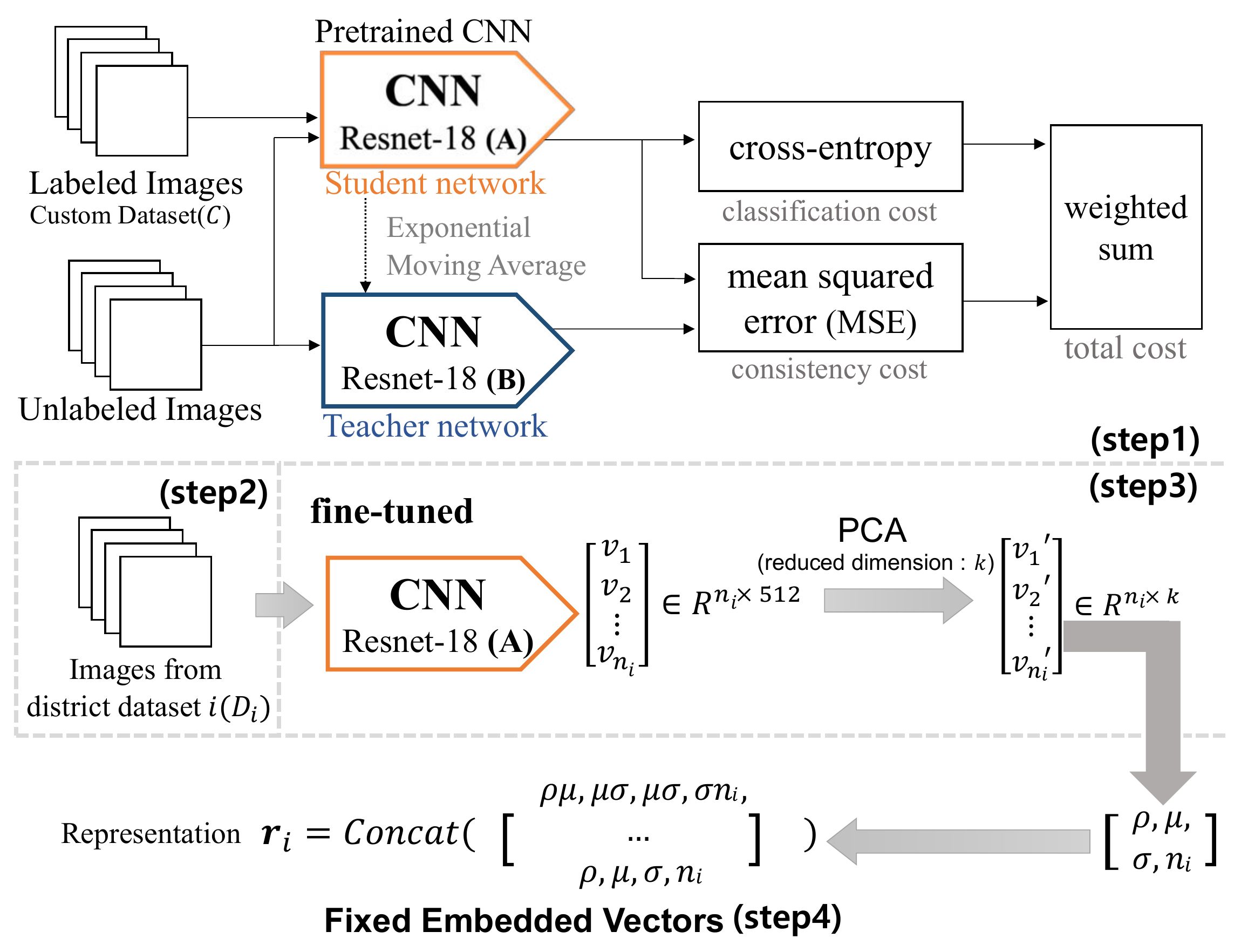}
    \caption{Our model operates in four steps: (step1) training embedding via semi-supervised learning and transfer learning, (step2) data pruning, (step3) dimensionality reduction, and (step4) calculating the embedded spatial statistics and conducting regression for validation.
    }
    \label{fig:modeling1}
\end{figure}

We first state the problem. Let $d_j^i$ be the $j$-th satellite imagery of district $i \in U$, where $U$ is the complete set of districts in a country. Let $D_i$ be the set of satellite imagery of district $i$. Since districts can be of any shape and size, the number of satellite images in $D_i$ varies from one district to another. We define this number for district $i$ as $n_i$, i.e., $d_j^i \in D_i$ where $1 \leq j \leq n_i$. Then, the main problem is defined as follows: 
\begin{quote}
\textbf{Problem:~}
{Given an image set $D_i$ of district $i$, can we extract fixed-sized ($s$) representations $\mathbf{r}_i$ (i.e., $\mathbf{r}_i \in \mathbf{R}^s$ of any district $i$) and predict $y_i$ the attribute of interest in the district $i$}? 
\end{quote}
In Fig.~\ref{fig:sate_demo2}, this would be equivalent to extracting a representation and predicting the related attributes of a district given its complete set of satellite image tiles. 

\begin{figure*}[t!]
    \centering
    \includegraphics[width=2.1\columnwidth]{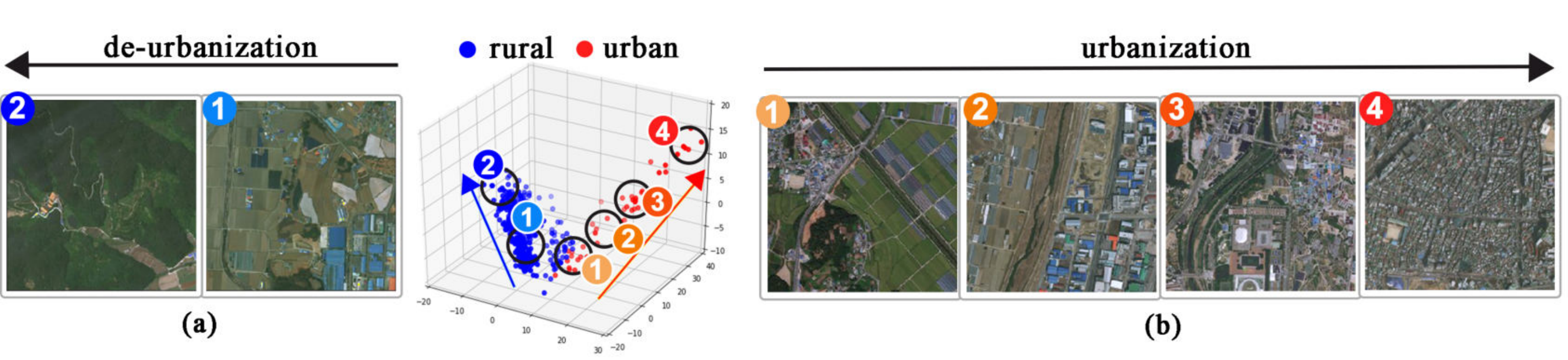}
    \caption{Embedded space analysis shows that rural images (blue) are well separated from more urban images (red). For the blue cluster, as we observe images from  anchor points 1 to 2, the de-urbanization trend becomes more pronounced. In contrast, for the red cluster, as we observe images from anchor points 1 to 4, the degree of urbanization becomes more intense. 
    }
    \label{fig:urbanize}
\end{figure*}

\subsection{Training embedding via semi-supervised learning and transfer learning}
Network learning with supervision was used to extract meaningful information from satellite imagery. We constructed a labeled custom dataset ($\mathcal{C}$) that includes 1,000 randomly selected satellite images and employed the following three labels, which are directly related to economic scales: urban, rural, and uninhabited. We hired four annotators to obtain the labels of the images. We integrated the decisions of all annotators as soft labels (i.e., average votes), which were then used to build a classifier that divides satellite images into three classes. However, obtaining reliable labels for each satellite image tile was a time-consuming task. Here, a key challenge was the relatively small number of labeled data, which was addressed by adapting a semi-supervised learning approach.

Semi-supervised learning trains classifiers baesed on a combination of a small amount of labeled data and a large amount of unlabeled data. Mean Teacher~\cite{tarvainen2017mean}, which is a powerful model in this domain, utilizes unlabeled data to penalize predictions that are inconsistent between the student and teacher models. This regularization technique can provide smoothing in the decision boundary for a robust and accurate forecast. We used the Mean Teacher architecture with the ResNet18 backbone for training, where the loss for a labeled and unlabeled dataset is as follows:
\begin{align*}
&\mathcal{L}_{labeled} =  -\sum_{i \in label}{\Tilde{y}_{i}\log f({d})_{i}} \\
&\mathcal{L}_{unlabeled} =  \| f({d}) - \Tilde{f}({d}) \|_2^2 \\
&\mathcal{L}_{total} = \mathcal{L}_{labeled} + w(t) \times \mathcal{L}_{unlabeled}
\end{align*}
, where $\Tilde{y}$ is the ground truth class probability (i.e., rural, urban, and uninhabited), $f$ is the student mode we aim to optimize, and $\Tilde{f}$ is the teacher model whose weight follows the exponential moving average of the student model. The total loss is the sum of two losses, and we increase the weight of the unlabeled loss from 0 to 12.5 during the first 40 epochs.

Transfer learning is another powerful approach to overcoming small data. The knowledge transferred from a similar but larger labeled dataset can efficiently help tune a model with a small labeled dataset~\cite{yosinski2014transferable,oliver2018realistic}. In addition to semi-supervised learning, we concurrently adopted transfer learning by first pretraining the CNN model with the ImageNet dataset~\cite{deng2009imagenet} and then using the pretrained model as an initial student network in the Mean Teacher model.

We used 1,000-sample labeled data and 22,577-example unlabeled data and apportioned the labeled data into the train and test sets by an 80-20 split. The model achieved a 92.0\% accuracy when evaluated based on the majority-voted label of the test set. Then, the fine-tuned model was used as the feature extractor, replacing the final fully connected layer. Images with large dimensions ($d_j^i \in \mathbf{R}^{256 \times 256 \times 3}$, 256x256 pixels with an RGB band) can be reduced to relatively low dimensional vectors ($v_j^i \in \mathbf{R}^{512}$, where 512 is the size of the final layer in ResNet18 excluding the fully connected layer). 

To determine whether the classifier extracts essential features, we visualized sample images into three-dimensional space by reducing the embedded vectors by PCA. Fig.~\ref{fig:urbanize} displays the extracted features in the reduced vector space of sample images of various urban and rural areas. Here, the rural and urban images are separated and aligned well in the virtual direction (i.e., red and blue arrows). Furthermore, these virtual axes represent the degree of urbanization. The left-hand side of the picture shows two satellite image tiles. Both tiles have rural characteristics, and the images that seem to contain a smaller human population are positioned further toward the blue arrow (i.e., tile ``2'' seems less urbanized than tile ``1''). The image tiles on the right-hand side contain capture more populated areas. Tiles ``3'' and ``4'' that are toward the end of the red arrow show a highly urbanized cityscape, whereas tiles ``1'' and ``2'' contain fewer residential areas. This figure demonstrates the strength of our model in its ability to learn high-level features and align satellite images along these virtual axes.

\subsection{Data pruning}
According to the Global Rural-Urban Mapping Project, only 3\% of the land cover is an urban area, and approximately 40\% of the land over is an agricultural area~\cite{doxsey2015taking,foley2005global}. The remaining uninhabited area accounts for the largest portion of the earth. Since such regions do not show human artifacts, they could act as noise when extracting representations related to human activities. We built a CNN classifier by filtering areas that are probably uninhabited. For the training, we reused a custom dataset that included 1,000 randomly selected satellite images. Three annotators labeled the images as either inhabited or uninhabited, and their majority votes were used as the ground truth labels. This process labeled 53.9\% of the images as inhabited, and the remaining 46.1\% of the images were labeled uninhabited.

The CNN classifier included the pretrained ResNet18 model with the ImageNet dataset. The dataset was randomly shuffled and split into a training set (80\% of the data) and a test set (20\%). The model was trained to minimize cross-entropy loss for binary classification. Stochastic gradient descent was used to reduce the loss term, and data-augmentation methods, such as rotating or flipping figures, were used to increase the amount of training data. As a result, the classifier achieved 95.5\% accuracy in the test set. We removed all images that were labeled uninhabited from each district. Of the initial 96,131 images, 51,618 (53.702\%) images were removed in this manner.

\subsection{Dimensionality reduction of embedding}
The next step reduced the 512 dimensions from the final layer in ResNet18 into smaller sizes. Since our goal is to predict attributes of interest $y_i$ and obtain a unique representation from a pruned image set $\hat{D_i}$ across districts (i.e., $n$=230 administrative districts), we aimed to produce a dimension size $v_i$ smaller than the number of districts $n$ to avoid overfitting. We implemented a principal component analysis (PCA) to reduce the dimensions of the embedded features $v_i$, which appears in the center of Fig.~\ref{fig:modeling1}. 

PCA is nonparametric and does not require a parameter tuning process. PCA uses orthogonal linear transformations of the original vectors to extract principal components with the maximum variance. A sufficient number of principal components should explain most of the variance in the data while efficiently reducing dimensions. To determine this number ($k$), we investigated how each new principal vector explained much variance. The trend shows that the first four components explain approximately 80\% of the variance and that additional gains rapidly become marginal. After the 10th component, the gain is less than 0.5\% of the total variance. We consider up to the first ten principal components for the dimensionality reduction, i.e., $k$ $(1 \leq k \leq 10)$.

\subsection{Presenting the embedded spatial statistics}

This final step addresses the challenge arising from the varying input size in which a different number of image tiles define districts. Previous studies in a different domain have attempted to address such arbitrary input length problems via preprocessing techniques, such as adding sequence padding or recurrent neural network-based learning~\cite{yang2016hierarchical,hochreiter1997long}. However, these methods cannot resolve the substantial differences in input lengths that are typical in demographic research. The smallest district could be covered by fewer than ten image tiles, whereas the largest district requires more than hundreds of tiles, resulting in orders of magnitude difference. 

We present a technique to summarize any length of image features into a fixed set of vectors. Let $g$ be the composition of the fine tuned feature extractor and $k$ $(1 \leq k \leq 10)$ be the resulting principal components. All images $d_j^i$ in $\hat{D_i}$ are transformed to $v_j' \in \mathbf{R}^{k}$ by $g$. Let the matrix of the final embedded vectors from district $i$ be $R_i \in \mathbf{R}^{n_i \times k}$. 

To produce a fixed-length embedding from vast geographic areas, we propose to utilize the following descriptive statistics: (i) the mean $\mu$, (ii) the standard deviation $\sigma$, (iii) the number of satellite images of a district $n$, and (iv) Pearson's correlation of the dimensionally reduced features $\rho$. These four quantities are fundamental embedded spatial statistics capturing the observation that satellite images of areas with geo-proximity exhibit similar traits. Descriptive statistics represent data by central tendency (mean, median, and mode), dispersion (variance, standard deviation, and skewness), and association (chi-square and correlation). The proposed quantities are descriptive statistics representing satellite images that belong to the same district.

\begin{figure}[ht!]
    \centerline{
    \includegraphics[width=1\columnwidth]{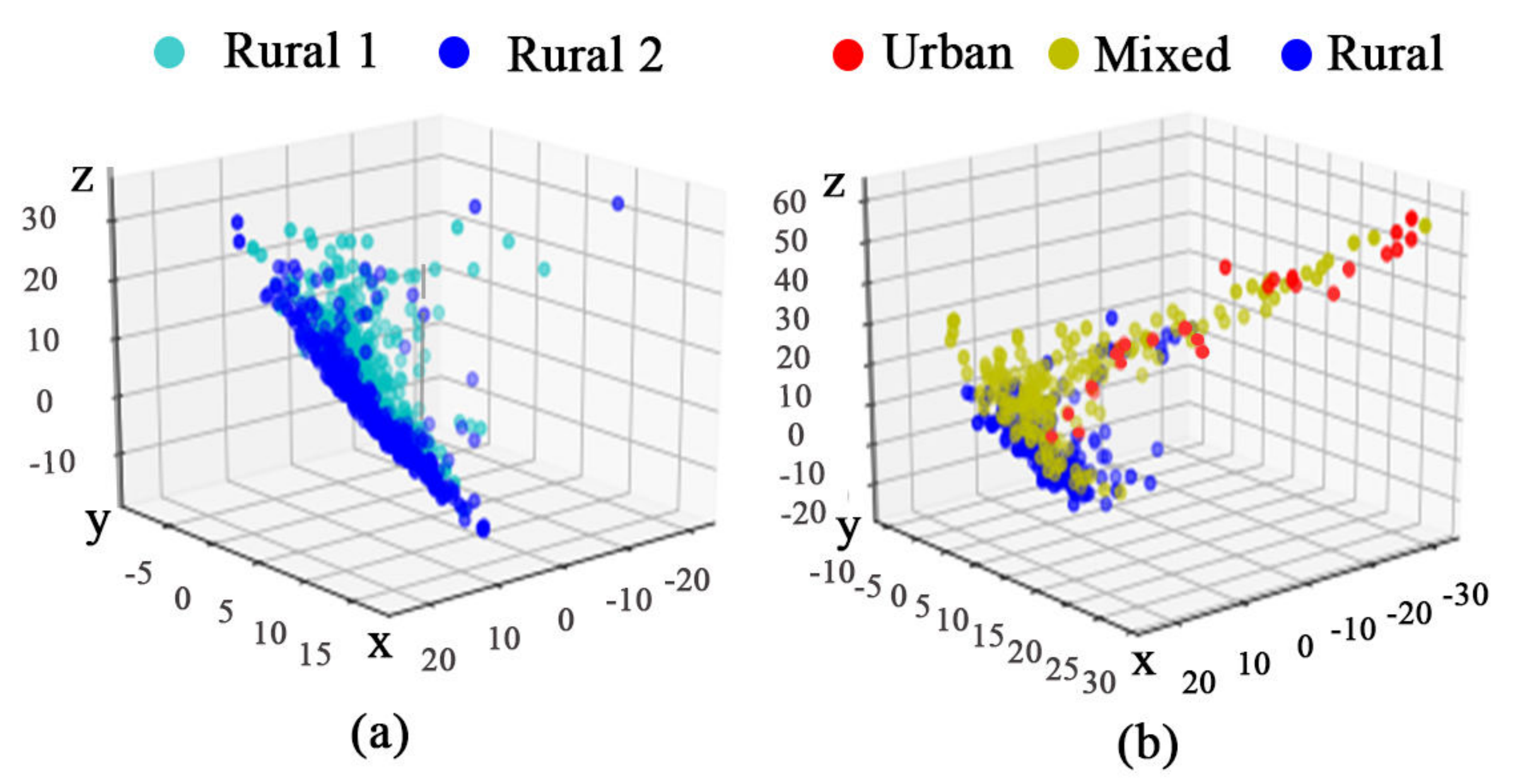}}
    \centering
    \caption{Embedding examples: (a) two rural areas show similar traits on the reduced dimensions; (b) semi-urban areas (marked as mixed) possess both traits of rural and urban areas.}
    \label{fig:example} 
\end{figure}

Fig.~\ref{fig:example} displays the feature vectors of the satellite images of five districts over a 3-dimensional space. Two rural districts appearing on the left-hand side exhibit a similar embedding space, whereas the three districts on the right-hand side exhibit different embedding spaces. Densely urbanized areas (red dots) have high variances ($y$-axis) and high mean values ($z$-axis) in the figure. In contrast, vast rural areas, which are denoted by blue or light blue dots, have comparatively low variances and low mean values ($z$-axis). The images on the right-hand side were chosen from a new district town that has a mixture of rural and urban areas and reports its embedded space in `mixed' (yellow dots). The satellite images belonging to this mixed district indeed possess statistical characteristics intermediate between urban and rural. Similar to these examples, areas with distinct characteristics have different statistical values; thus, the summarized features can reflect the characteristics of districts. 

Finally, cross-products of features were added to consider interactions to enrich the information regarding unknown embedded space distributions. These complete sets of features were learned per district $i$, as illustrated in the bottom part of Fig.~\ref{fig:modeling1} and became a fixed-sized representation $\mathbf{r}_i$. To predict the $y_i$ value for district $i$, we used $\mathbf{r}_i$ to fit a regressor.

\section{Data}
This study utilizes the following data: regional-level demographics and high-resolution satellite imagery. Both data types from many developed countries are accessible via the REST APIs of Esri\textregistered ArcGIS, a famous repository of maps and geographic~\cite{johnston2001using}. We chose South Korea as a representative developed country for training the model. Then, among all available satellite images of South Korea, we further identified those in which at least three vertices of an image tile belong to the polygon representing the boundaries of each district, as illustrated in Fig.~\ref{fig:sate_demo2}. This heuristic is simple but reasonable for addressing various polygon shapes. In total, 96,131 satellite images (256x256 pixels) of 230 South Korean districts were collected in this manner. The utilization of all image tiles per district distinguishes our work from those of others, c.f., previous studies used a fixed set of satellite images; -- for example, a seminal study conducted in African countries used 100 randomly chosen images tiles over 10x10$km^2$ areas~\cite{jean2016combining}.

\subsection{Satellite imagery data}
The World Imagery satellite data captured by DigitalGlobe provides 256x256-pixel image tiles over a wide range of zoom levels, $Z$ (0$\sim$18). While a single increment in the zoom level enlarges the area within both the vertical and horizontal directions, the overall number of pixels maintains the same size. Hence, the resolution becomes double, and the coverage becomes quarter per increment. 

The highest freely available resolution is 0.6m-resolution, which is $Z$=18. However, under such resolution, image annotation becomes challenging because of the small spatial coverage of each image tile. Moreover, the size of district data increases exponentially, leading to computation overhead in training the model. In contrast, choosing a zoom level below 15 results in poor performance because one image tile may contain multiple districts. The smallest district in data spans 2.8$km^{2}$, which cannot be distinguished at a zoom level 14. We chose the zoom level of 15 with 4.7m-resolution, which allows for the identification of large objects that cover more than 5x5$m^2$. For example, we can start to recognize buildings and roads but not the cars on the street at $Z$=15. Each image covers approximately 1.4$km^2$, which is considered appropriate for observing human settlements in the previous research~\cite{jean2016combining} using a satellite image of coverage 1$km^2$. 


\begin{table}[t!]
\caption{Example of demographic attributes}
\label{adv_demographics}
\centering
\resizebox{1\columnwidth}{!}{
\begin{tabular}{lllrrrr}
\hline
\multicolumn{1}{l}{\textbf{Category}} & \multicolumn{1}{l}{\textbf{Variable ID}} & \multicolumn{1}{l}{\textbf{Description}} \\
\hline
Population & density & Population per square kilometer\\ 
Age  & 0-14 & Population density by age 0-14\\ 
& 15-29 & Population density by age 15-29\\
& 30-44 & Population density by age 30-44\\
& 45-59 & Population density by age 45-59\\
& 60+ & Population density by age 60+\\
Education  & nodegree & Population density by no degree\\ 
& elementary & Population density by elementary school\\
& middle & Population density by middle school\\
&\multicolumn{1}{c}{...} & \multicolumn{1}{c}{...}\\ 
& phd & 
Population density by Ph.D. degree\\
Household & count & Household count per square kilometer\\ 
& size & Average household size\\
Income & total & Total purchasing power per household\\ 
& capita & Purchasing power per capita \\
\hline
\end{tabular}
}
\end{table} 

\subsection{Demographics data}

The Esri Demographics by Michael Bauer Research GmbH provides 2018 demographics data and the boundary polygon shapes of districts in 135 countries. For South Korea, the following data types were available: Population, Age, Education, Household, and Income. The population density and 27 other demographics measured per capita were used as representative ground truth data in this paper. Table~\ref{adv_demographics} displays example demographic fields and their descriptions.

\section{Results}
\subsection{Performance evaluation and ablation study}

We conduct a set of experiments. The first evaluation takes advantage of the population demographics by dividing these data into a training set and a test set in an 80\%–20\% ratio. 4-fold cross-validation is applied to the training data set to tune the model's hyperparameters, such as the PCA dimensions and the regularization term in the cost function.

We implemented nine baselines to evaluate. \textbf{Nightlight} uses the total light intensity of the districts from nighttime satellite imagery to predict economic scales~\cite{bagan2015analysis}. A regressor was built and trained to obtain the sum of nightlights in each district. Then, \textbf{Auto-Encoder} extracts compact features as step2 on our model. An autoencoder is an unsupervised deep learning algorithm that does not need any label information. The model aims to learn an approximate identity function to construct an output that is similar to the input while limiting the number of hidden layers. \textbf{No-Proxy} is identical to the proposed model but lacks any knowledge transfer from the proxy dataset. This model was pretrained only with the ImageNet dataset and, hence, can demonstrate the value of the custom dataset. 

To verify the effectiveness of READ compared to a well-known model~\cite{xie2016transfer}, we trained \textbf{JMOP} (Jean Model with Our Proxy) which is a combination of two models. First, we use a proxy which is predicting for rural, urban, and inhabited classes in the same method of READ. Then, we summarize the features and use them to predict with an identical set of model~\cite{xie2016transfer}. Finally, \textbf{SOTA} is the best known grid-based approach for population density prediction~\cite{facebook2019data}. The implementation details of this model are not published, but the prediction results on each arc-second block (approximately 30x30$m^2$) are shared online. We could aggregate the published grid-level data across districts and regress such data with ground truth statistics. The four remaining baselines are ablation studies that remove each feature from READ.

\begin{table}[ht!]
\small
\centering
\caption{Baseline results and ablation study. Performance comparison on predicting population density by READ and nine baselines.}
\label{baseline_results}
\resizebox{0.9\columnwidth}{!}{
\begin{tabular}{l c l}
\hline
\multicolumn{1}{c}{\textbf{Model}} & \multicolumn{1}{c}{\textbf{MSE}} & \multicolumn{1}{c}{\textbf{R-Squared}} \\
\hline 
Nightlight & 0.4254$\pm$0.0664 & 0.6133$\pm$0.0635\\
Auto-Encoder & 1.6242$\pm$0.3445 & 0.6347$\pm$0.0823\\
No-Proxy    & 0.2800$\pm$0.1118 & 0.7359$\pm$0.1117\\
JMOP    & 0.4448$\pm$0.0998 & 0.8985$\pm$0.0253\\
SOTA &  -  &    \multicolumn{1}{l}{\textbf{0.9231}}  \\
\hline
READ w/o $\mu$     & 0.2612$\pm$0.0632 & 0.9429$\pm$0.0155\\
READ w/o $\rho$    & 0.2165$\pm$0.0596 & 0.9527$\pm$0.0140\\
READ w/o $n$       & 0.1921$\pm$0.0471 & 0.9579$\pm$0.0119\\
READ w/o $\sigma$  & 0.1902$\pm$0.0592 & 0.9586$\pm$0.0130\\
READ    & \textbf{0.1761$\pm$0.0383} & \textbf{0.9617$\pm$0.0090}\\
\hline
\end{tabular}}
\end{table}

All models were trained with an 80\%-20\% train-test ratio and 4-fold cross-validation. XGBoost~\cite{chen2016xgboost} was used to enhance prediction accuracy. The models were evaluated 20 times with a randomly split dataset. Table~\ref{baseline_results} reports the mean and standard deviation of the prediction performances. READ outperforms all of the nine baselines in both the R-squared ($R^2$) and mean squared error (MSE) values. Our model even outperforms the current state-of-the-art (SOTA) approach, which is~\cite{facebook2019data}. We find that transfer learning from the custom labeled dataset helps produce a more meaningful embedded, by distilling knowledge associated with urban and rural classifications. This finding is demonstrated by the increased prediction quality against two models: No-Proxy and Auto-Encoder. Furthermore, the quality gain over JMOP indicates that the summarizing technique of READ contributes massively to the performance gain. The ablation study shows that removing any of the descriptive statistics lowered the performance, indicating that $n$, $\mu$, $\rho$, and $\sigma$ all make a meaningful contribution.

\subsection{Comparisons of architecture alternatives}

Subsequently, we tested alternative design choices. First, we considered the following networks as the backbone CNNs: 1) DenseNet121, a densely connected convolutional network that connects each layer to each other layer in a feed-forward fashion~\cite{huang2017densely}; 2) AlexNet, an 8 layer network that won the ImageNet challenge 2012~\cite{NIPS2012_4824}; and 3) VGG16, a 16 layer very deep convolutional network~\cite{simonyan2014very}. These options were compared to our choice of ResNet18~\cite{he2016deep}. Using ResNet18 as the backbone CNN was found to yield the highest $R^2$ value, which is demonstrated in Table~\ref{backbone_results}. The difference, however, remains in the range between 0.021 and 0.0253. 

\begin{figure}[ht!]
    \centerline{
    \includegraphics[width=1\columnwidth]{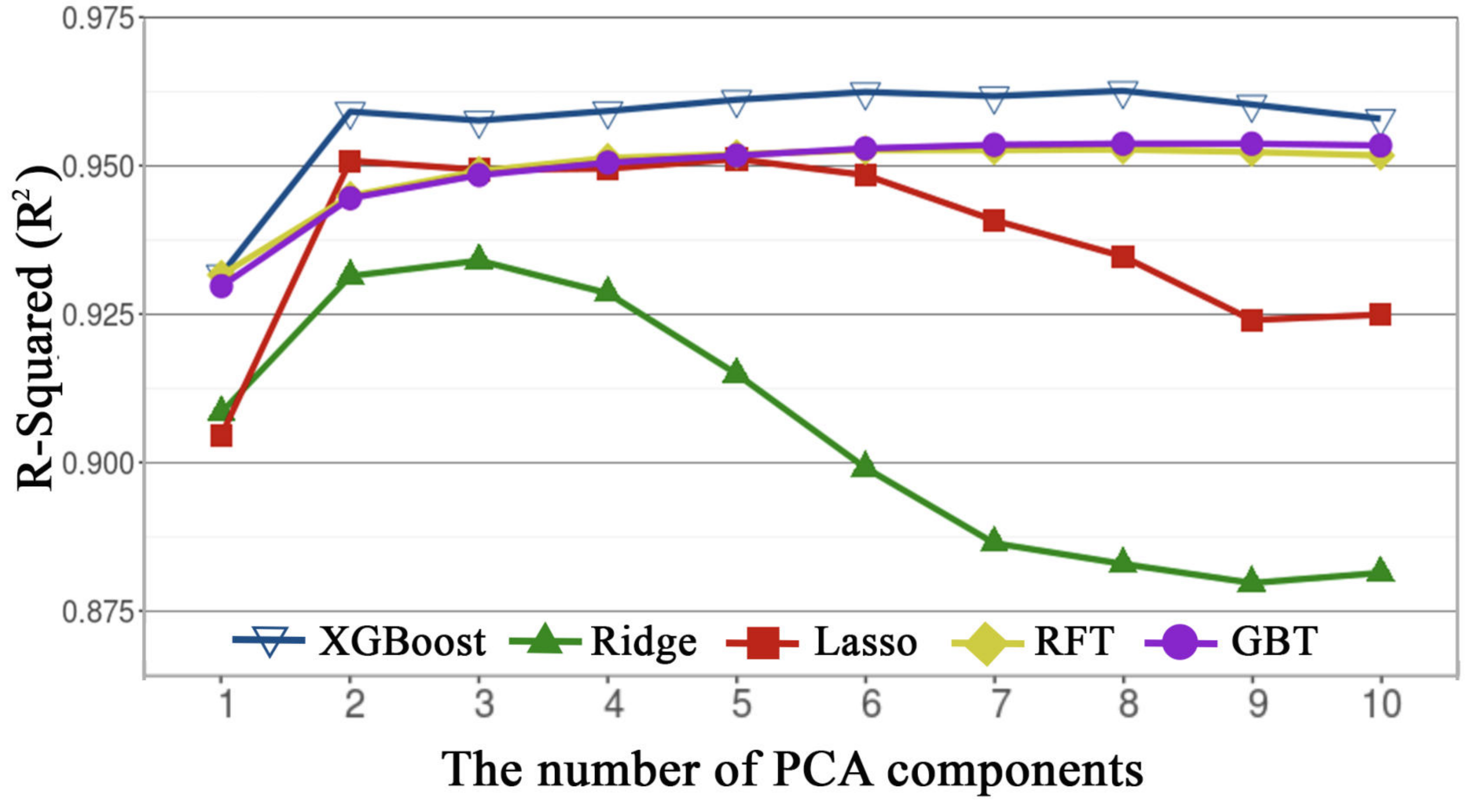}}
    \centering
    \caption{Prediction performances by the number of PCA components. } 
    \label{fig:comparison}
\end{figure}

\begin{table}[ht!]
\small
\caption{Comparison across different backbone networks.}
\label{backbone_results}
\centering
\resizebox{0.95\columnwidth}{!}{
\begin{tabular}{lcccc}
\hline
\multicolumn{1}{l}{\textbf{Model}} & \multicolumn{1}{l}{ResNet18} & \multicolumn{1}{l}{DenseNet121} &\multicolumn{1}{l}{AlexNet}&\multicolumn{1}{l}{VGG16} \\
\hline
$R^2$ & 0.9617 & 0.9337 & 0.9364 & 0.9407\\
\hline
\end{tabular} 
}
\end{table}
We also apply the following regressors: 1) XGBoost~\cite{chen2016xgboost}; 2) Ridge and Lasso as linear models with different regularization terms; 3) Random Forest (RFT) as a widely used ensemble model; and 4) Gradient Boosting Tree (GBT) as an ensemble model with boosting that nullifies overfitting. Fig.~\ref{fig:comparison} shows the prediction accuracy as a function of the dimension size. The Lasso and Ridge regression showed a decreasing performance over the component count after four components in the case of Ridge and six components in the case of Lasso regression. These results are likely due to high dimensional input features and the lack of effective regularization in those models. In contrast to these methods, XGBoost, Random Forest, and Gradient Boosting Tree were resistant to overfitting, and XGBoost achieved the best performance in our model.

\subsection{Evaluation over broader scales and countries}

The final evaluation reports predictions of a complete set of economic scales by READ. All values are log-scaled, and XGBoost is used. The average $R^2$ of 20 trials of prediction of the study area is shown in Table~\ref{total_result}. The third column represents READ applied to South Korea to predict the population density and its subclass divided by age very precisely ($R^2$ > 0.95). In the income category, the purchasing power per capita (capita) is predicted by $R^2$ as 0.7603. Finally, two demographics in the household category show an extreme difference in their prediction quality. While the household count per square kilometer (count) reports the highest $R^2$ of 0.9664, the average household size (size) reports the lowest $R^2$, i.e., 0.6181, among the economic scales.


\begin{table}[ht!]
\caption{Prediction performances of economic scales for South Korea and Vietnam.}
\label{total_result}
\centering
\resizebox{0.99\columnwidth}{!}{
\begin{tabular}{llccc}
\hline
\multicolumn{1}{l}{\textbf{Category}}{} & \multicolumn{1}{l}{\textbf{Variable ID}} & \multicolumn{1}{c}{\textbf{South Korea}} & \multicolumn{1}{c}{\textbf{Vietnam}} \\
\hline
Population & density & 0.9617 & 0.8863\\ 
Age & 0-14 & 0.9520 & 0.8756 \\ 
 & 15-29 & 0.9570 & 0.8791\\ 
 & 30-44 & 0.9575 & 0.8881\\ 
 & 45-59 & 0.9624 & 0.8804\\ 
 & 60+ & 0.9654 & 0.8731 \\ 
Household & count & 0.9664 & 0.8896\\ 
 & size & 0.6181 & 0.4460\\ 
Income & capita & 0.7603 & 0.6822 \\
\hline 
\end{tabular}
}
\end{table}

The high prediction capability of READ may be due to the custom dataset, which was built from the same country (see step1 in Fig.~\ref{fig:modeling1}). To test its applicability, we gathered a total of 226,305 satellite images along with economic scale data from another country, Vietnam. Then, we applied the model learned from South Korea to predict the economic scales of Vietnam. Surprisingly, READ achieved a high $R^2$ value of 0.8863 in predicting the population density in Vietnam, even when it was trained solely on data from a different country.  

The above finding demonstrates that the learned spatial representation of READ successfully captures general indicators of socioeconomic scales that extend beyond a single country use. It may be that Asian countries exhibit similar pathways in economic growth and demographic transition~\cite{vietnam_korea}. Interestingly, we also note a substantial difference in the prediction quality of specific demographic attributes between the two countries. For example, the highest $R^2$ value in the age category is found in the `60+' group in South Korea (0.9654) but in the `30-44' group in Vietnam (0.8881), which might have been affected by the concentration of the elderly population in rural areas in South Korea due to demographic transition and rural-urban migration. In contrast, a more substantial younger population is known to live in rural areas in Vietnam~\cite{UN_data}, compared to South Korea.

Finally, as another way to quantify how much gain building a custom labeled dataset of land covers contribute to the model, we conducted another experiment. We additionally built a custom labeled dataset for Vietnam by hiring two local annotators and obtained the land cover labels for 1,000 randomly selected images. We used this customized dataset to retrain the entire model. However, building a customized dataset for Vietnam only led to a marginal gain in predictions. The prediction of population density improved to the $R^2$ value of 0.8876. We expect that the minimal performance gain from building a custom dataset for Vietnam data is due to the quality of satellite images. Unlike in the case of South Korea, many satellite images of the region available from ArcGIS contained clouds, even after applying a cloud filter. 

\section{Discussion \& Conclusion}
This paper proposed a novel model that efficiently limits the number of dimensions via transfer learning to summarize any number of satellite images. The final spatial representation, which is a fixed-length embedded vector, can be used to estimate the socioeconomic growth of urban and rural areas at the district level. This method could benefit developing countries lacking the much-needed infrastructure to monitor their rapid urbanization process closely. Utilizing all of the high-resolution daytime satellite images, without restraining the model to handle random samples as in previous work, led to the substantial performance gain. Our approach can be widely applied by national and regional governments to estimate economic scales, such as the age and purchasing of district-level populations. Below, we discuss two potential application domains of READ.

\subsection{Revealing sub-district level scales}
\begin{figure}[ht!]
    \centerline{
    \includegraphics[width=1\columnwidth]{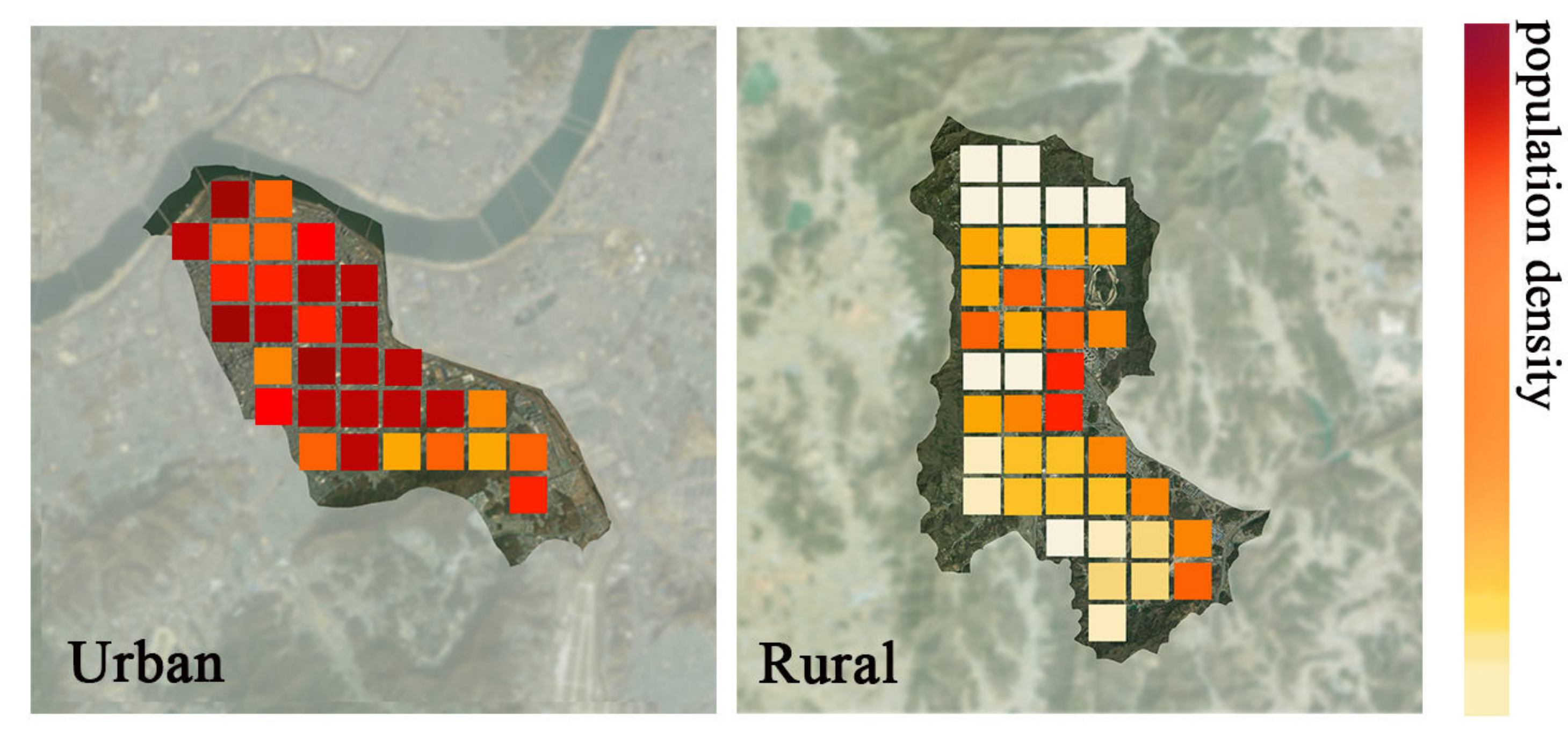}}
    \centering
    \caption{
    A single district often comprises multiple neighborhoods of different economic scales. READ naturally models such small-scale variations.
    }
    \label{fig:heatmap}
\end{figure}

While we present the results of district-level predictions, the proposed method predicts attributes over smaller levels, i.e., satellite image tiles. The potential to utilize such micro-level inference is magnificent. READ not only distinguishes urban and rural areas, as demonstrated in Fig.\ref{fig:heatmap}, but also captures the fine-grained layout of land cover. The rural area depicted in the figure indicates that the degree of urbanization is not even across the district and that the regions illustrated in white color, in particular, lack urban infrastructure. Such fine-grained demographic information could be highly valuable in helping answer questions, such as ``which areas of the city are growing the fastest or the slowest?'' and ``which areas outside the city lack critical infrastructure, such as roads?'' that could have been answered, otherwise, via expensive offline surveys. Our technique, READ, can sufficiently summarize the high-resolution satellite imagery to benefit neighborhood-level demographic research.

\subsection{Application to measure urban sprawl}

Cities are overgrowing worldwide, and this urbanization process is sometimes faster than the development of the necessary infrastructure needed for the urban population. This problem of tracking rapid urban growth, which is called urban sprawl~\cite{sprawl}, has been challenging due to the lack of data, particularly in rapidly evolving developing countries. We show with an example that applying READ over historical data can help better understand urban sprawls.

\begin{figure}[ht!]
    \centering
    \includegraphics[width=1\columnwidth]{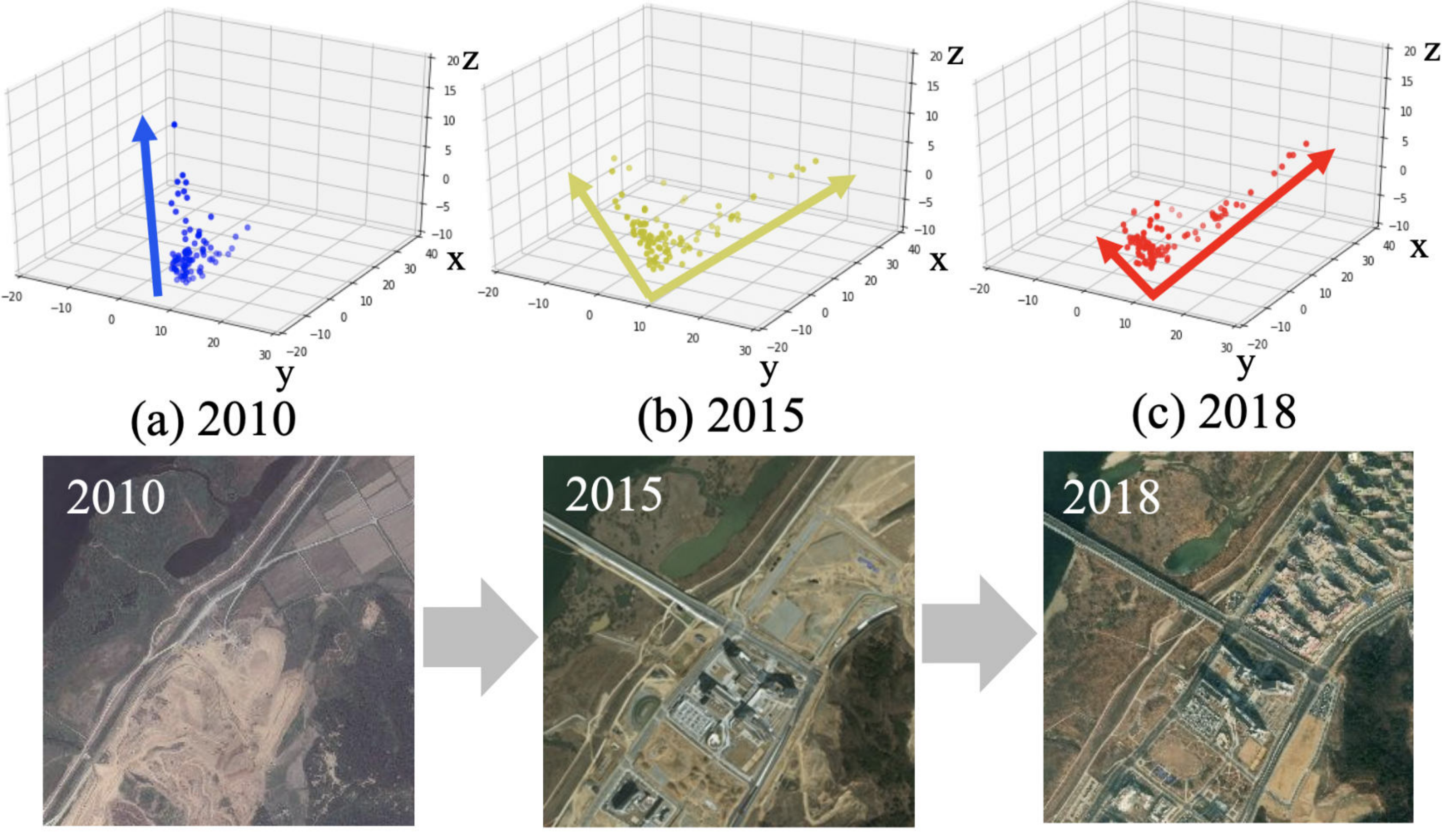}
    \caption{READ can be used to compare the evolution of the same district over multiple years. This given district exhibited mostly rural features in 2010, but its visual composition changed to contain a mix of urban and rural features towards 2018.
    }
    \label{fig:comparison_city2}
\end{figure}

Fig.~\ref{fig:comparison_city2} displays the embedded feature vectors of satellite images on the reduced dimensions of one district at different times over a 3-dimensional space. By comparing the embedded feature vectors of the same region in (a) 2010, (b) 2015, and (c) 2018, we observe different types of vector space. The densely urbanized area (c) is diffusing in the $x$-axis direction. However, the less developed area (a) is spreading along the $z$-axis. Our finding indicates the change in a particular area that is relatively fast-growing over time. Detecting the changes in a specific region is recognized as an essential aspect of implementing policy to improve the quality of the local economic activity.

\subsection{Limitations \& Future work}

This work has several limitations. First, the application of our model will be affected by the urbanization patterns and the quality of satellite imagery. Although we confirmed a high $R^2$ value in Vietnam for the model trained from South Korea data, such high applicability may be due to both countries showing similar patterns of urbanization, particularly in terms of building shapes~\cite{iimi2005urbanization}. The architectural similarity may have contributed to the consistent performance over a new country, even though the model was solely trained using another country's data. Nonetheless, we emphasize that land cover labels considered in the current paper (i.e., rural, urban, and uninhabited) are general. The model, therefore, is not specific to describing particular regional characteristics such as agriculture or mining. In the future, we plan to consider other benchmark data from vastly different cultural backgrounds. Second, our model utilizes only freely available data. Currently, rich geospatial data describing the elevation of surfaces such as buildings, vegetation, and other human artifacts, are becoming available, e.g., the Digital Surface Model (DSM) data. While DSM data is costly to obtain, future studies may incorporate alternative datasets to advance characterizing urban agglomerations further. Finally, our approach lacks interpretability. Due to the black-box nature of the deep learning model utilized, the model cannot determine which image feature most contributes to urbanization. We aim to enhance interpretability by adopting explainable structures, such as grad-CAM~\cite{selvaraju2017grad}.

\section{Acknowledgments}
We sincerely thank Hyunjoo Yang, Jihee Kim, and Sangyoon Park for their valuable feedback on this work. \\This research was supported by the Basic Science Research Program through the National Research Foundation funded by the Ministry of Science and ICT in Korea (No. NRF-2017R1E1A1A01076400).

{
\small
\bibliographystyle{aaai}
\bibliography{aaai20_main}
}

\appendix
\end{document}